# Interpolating Conditional Density Trees

Scott Davies and Andrew Moore
School of Computer Science
Carnegie Mellon University
Pittsburgh, PA 15213
[scottd, awm]@cs.cmu.edu

## Abstract

Joint distributions over many variables are frequently modeled by decomposing them into products of simpler, lower-dimensional conditional distributions, such as in sparsely connected Bayesian networks. However, automatically learning such models can be very computationally expensive when there are many datapoints and many continuous variables with complex nonlinear relationships, particularly when no good ways of decomposing the joint distribution are known *a priori*. In such situations, previous research has generally focused on the use of discretization techniques in which each continuous variable has a single discretization that is used throughout the entire network.

In this paper, we present and compare a wide variety of tree-based algorithms for learning and evaluating conditional density estimates over continuous variables. These trees can be thought of as discretizations that vary according to the particular interactions being modeled; however, the density within a given leaf of the tree need not be assumed constant, and we show that such nonuniform leaf densities lead to more accurate density estimation. We have developed Bayesian network structure-learning algorithms that employ these tree-based conditional density representations, and we show that they can be used to practically learn complex joint probability models over dozens of continuous variables from thousands of datapoints. We focus on finding models that are simultaneously accurate, fast to learn, and fast to evaluate once they are learned.

## 1 INTRODUCTION

Bayesian networks are a popular method for representing joint probability distributions over many variables. A Bayesian network contains a directed acyclic graph $G$ with one vertex $V_i$ in the graph for each variable $X_i$ in the domain. The directed edges in the graph specify a set of independence relationships between the variables. Define $\vec{\Pi}_i$ to be the set of variables whose nodes in the graph are "parents" of $V_i$. The set of independence relationships specified by $G$ is then as follows: given the values of $\vec{\Pi}_i$ but no other information, $X_i$ is conditionally independent of all variables corresponding to nodes that are not $V_i$'s descendants in the graph. These independence relationships allows us to decompose the joint probability distribution $P(\vec{X})$ as $P(\vec{X}) = \prod_{i=1}^{N} P(X_i|\vec{\Pi}_i)$, where $N$ is the number of variables in the domain. Thus, if in addition to $G$ we also specify $P(X_i|\vec{\Pi}_i)$ for every variable $X_i$, then we have specified a valid probability distribution $P(\vec{X})$ over the entire domain.

Bayesian networks are most commonly used in situations where all the variables are discrete; if continuous variables are modeled at all, they are typically assumed to follow simple parametric distributions such as Gaussians (e.g. (Heckerman and Geiger, 1995)). Some researchers have recently investigated the use of complex continuous distributions within Bayesian networks; for example, weighted sums of Gaussians (Driver and Morrell, 1995), Gaussian kernel-based density estimators (Hofmann and Tresp, 1995), and Gaussian processes (Friedman and Nachman, 2000) have been used to approximate conditional probability density functions. Such complex distributions over continuous variables are usually quite computationally expensive to learn. This expense may not be too problematic if an appropriate Bayesian network structure is known beforehand. On the other hand, if the dependencies between variables are not known *a priori* and the structure must be learned from data, then the number of conditional distributions that must be learned and tested while a structure-learning algorithm searches for a good network can become unmanageably large.

In such cases, the search over network structures is usually performed using a discretized version of the data, where the range of each variable is divided into some number of bins and all values of a given variable within a given bin are considered equivalent. This discretization can performed once before network structure-learning, and the resulting network structure can then be reparameterized with continuous distributions in a final step ((Monti and Cooper, 1998b), (Monti and Cooper, 1999)); or, a simultaneous search of both network structures and discretization policies can be performed ((Friedman and Goldszmidt, 1996a), (Monti and Cooper, 1998a)). In this previous research, however, the discretization of each variable has been *global* – that is, the same discretization for any particular variable is employed for all the interactions in which it is involved.

Decision trees (see e.g. (Quinlan, 1986), (Breiman et al., 1984)) have been used previously in Bayesian networks over discrete variables (Friedman and Goldszmidt, 1996b) in cases where full conditional contingency tables could be too large to learn accurately from limited data. In this paper, we propose and evaluate four different tree-based approaches to the conditional density estimation of continuous variables, with different tradeoffs between accuracy, learning speed, and evaluation speed:

- CART(Breiman et al., 1984)-like trees, which are fast to learn and evaluate but are inadequate for accurately representing complex conditional distributions.

- *Stratified conditional density trees*, which are more computationally expensive to learn than CART-like trees but are still fast to evaluate and are better than CART-like trees at general-purpose density estimation.

- Joint density trees that are used conditionally. These are fast to learn, and (somewhat surprisingly) appear more accurate than stratified conditional density trees. Unfortunately, they are computationally expensive to evaluate.

- *Approximately conditionalized joint density trees*, which combine the best features of the previous three tree types in that they are fast to learn, fast to evaluate, and accurate at density estimation.

In Sections 2.1- 2.7, we explain, compare and contrast these four different types of trees, and provide experimental results on real and synthetic datasets in which we keep the set of conditional distributions being modeled constant. In Section 2.8, we briefly discuss a Bayesian network structure-learning algorithm that employs these trees, and show that the resulting overall algorithm can practically find accurate factored models that are also fast to evaluate, compared to global mixture model-learning algorithms such as AutoClass. Finally, in Section 3 we summarize our findings and discuss possible avenues for future research.

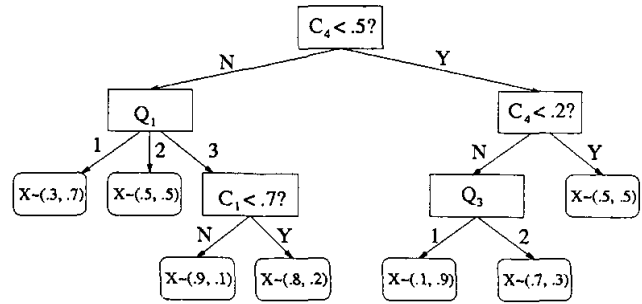

Figure 1: An example of a conditional density tree (or classification tree) for predicting the distribution of a binary variable $X$ as a function of several other variables.

## 2 TREE AND LEAF TYPES

### 2.1 Classification and regression trees

Figure 1 shows an example decision tree in which the distribution of a binary variable $X$ is predicted as a function of several other variables, some of which are discrete (the $Q$'s) and some of which are continuous (the $C$'s). To find the distribution of $X$, the prediction algorithm simply starts at the root of the tree (shown at the top of our diagram) and follows a path down the tree's branches according to the values of the other variables until it reaches a leaf. For example, if the continuous variable $C_4$ is less than .5, and the trinary discrete variable $Q_1$ has a value of 1, then the algorithm would predict that $X$ has a 30% chance of taking on its first possible value and a 70% chance of its second. Such decision trees for predicting the distributions of discrete variables are also known as "classification trees" in contexts where the task ultimately involves guessing a single value (typically the most likely value) for the variable being predicted.

Regression trees (e.g. (Breiman et al., 1984)) have structures similar to those of decision trees, but the leaves of these trees provide information about the distributions of a continuous variable $X$ instead. Typically in regression this information is restricted to a point estimate of the variable's mean; this mean may be constant, or it may be (for example) a linear function of the parent variables. In order to obtain an actual density estimate, a variance can be supplied as well as the mean in order to specify the parameters of a Gaussian.

Decision and regression trees are typically learned by greedy top-down divide-and-conquer algorithms; we employ such an algorithm in the experiments described in this paper. A decision or regression "stump" of depth one is grown for each possible branching variable. The algorithm then chooses the branch variable whose corresponding stump most increased the total conditional log-likelihoods of a randomly selected subset of the training data that was held out during the stump-training process. The algorithm then recursively learns the branch node's children using the appropriate subsets of the training data. When splitting on a discrete variable, the resulting branch always has one child for every possible value of the branch variable; when splitting on a continuous variable, the branch has two children corresponding to whether the variable is $\leq$ or $>$ the midpoint of the current possible subrange for that variable. (The algorithm is initially provided with a hypercube over the continuous variables in which all nonzero probability is assumed to lie.) Branching is terminated when fewer than ten training datapoints are consistent with the current subtree. A separate random holdout set of the training data is then used to prune the learned decision tree. Many variations of this learning algorithm are considered in the full version of this paper (Davies, 2002).

Regression trees may be adequate for representing continuous conditional distributions in situations where they are in fact near-Gaussian, or when the problem involves guessing a point estimate and then being penalized by its squared distance to the real value. However, there are other situations in which we may wish to have reasonably accurate models of distributions that are more complicated, e.g. multimodal. There are many possible criteria to use when judging the accuracy of such models; one of the most common is the Kullback-Leibler divergence of the model from the true distribution. Since we will be learning models from scientific data with unknown true distributions in our evaluations, we will use the log-likelihoods of test sets in cross-validation experiments as empirical analogues of the KL divergence.

## 2.2 Stratified conditional density trees

There is no reason in principle to stop at a simple parametric distribution for the child variable once the branching on parent (i.e. "input") variables has finished. Instead, one can employ a *stratified conditional*

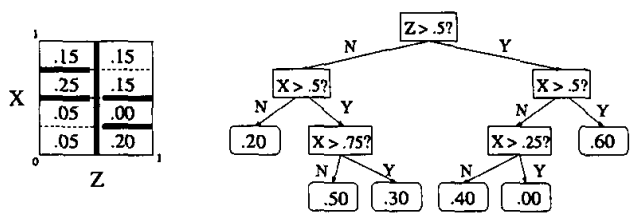

Figure 2: A stratified conditional density tree.

*density tree* in which any path from the root of the tree to a leaf first passes through a sequence of branch nodes that only test the parent variables, and then through another sequence of branch nodes that only test the child (i.e. "output") variables. Such a tree for estimating the conditional density of one continuous variable $X$ given another $Z$ might look like the one in Figure 2, where for clarity we have listed the conditional probability *masses* inside the leaves rather than the conditional probability densities; the densities are trivially computed from these masses by dividing by the volumes of the leaves. Note that in order to represent a valid conditional distribution, the masses in any subtree containing no branches on the parent variables must sum to 1. This constraint is what forces us to learn trees with this stratified branching structure: if branches on the input and output variables are allowed to alternate arbitrarily, then the constraint becomes nonlocal, making divide-and-conquer approaches to learning the tree difficult. (See (Davies, 2002) for details.) The recursive algorithm we employ to learn stratified conditional density trees is identical to the algorithm we use to learn decision and regression trees, except that wherever the decision or regression tree learner would call a routine that returns a simple leaf distribution fitting the provided data, the stratified conditional density tree learner calls a subtree-learning procedure. This subtree-learning procedure is also identical to the algorithm we use to learn decision and regression trees, except the subtrees *it* learns branch only on the output variable, and at each branch the algorithm divides the total conditional probability mass that each child is allocated according to the proportion of datapoints that fall in that child's subtree. Because an entire subtree is learned where a CART-like algorithm merely needs to learn a leaf distribution, learning stratified conditional density trees in this manner is significantly more computationally expensive to learn than CART-like trees. However, as we shall see, they can provide much more accurate density estimation.

Given the total conditional probability mass that lies a given leaf, we are still left with a choice of how to distribute it within that leaf. If the mass within each leaf is distributed with uniform density, then the stratified conditional density tree is essentially employing variable-resolution histograms in place of the simple parametric distributions (such as Gaussians) that CART-like trees use. However, most other choices (as mentioned in Section 2.5) lead to more accurate density estimation.

### 2.3 Using joint density trees conditionally

While the stratified conditional density trees discussed in the previous section can model conditional density trees much more accurately than CART-like single-level conditional density trees, they are computationally expensive to learn. There are many heuristics that could be tried to alleviate this problem, such as learning a CART-like tree first and using this tree's structure as a starting point for a stratified conditional density tree. However, such heuristics would be unlikely to increase the accuracy of the resulting models, and likely to decrease it. As it turns out, it is possible to achieve more accurate density estimation *and* faster learning using an alternative approach.

In this section we discuss the use of density trees modeling joint distributions $P(X_i, \vec{\Pi}_i)$ to obtain conditional density estimates $P(X_i|\vec{\Pi}_i)$. Each leaf $l$ of such a tree specifies a joint probability $P(X_i, \vec{\Pi}_i|l)$: that is, the probability that $X_i$ and $\vec{\Pi}_i$ take on specific values within the leaf's range, given that the datapoint lies somewhere within the bounds of $l$. Assuming we have a density tree representing $P(X_i, \vec{\Pi}_i)$, we can obtain an estimate for a particular $P(x_i|\vec{\pi}_i)$ as follows:

$$\begin{aligned}
P(x_i|\vec{\pi}_i) &= \sum_l P(l|\vec{\pi}_i) \cdot P(x_i|\vec{\pi}_i, l) \\
&= \sum_l \frac{P(l) \cdot P(\vec{\pi}_i|l)}{\sum_{l'} P(l') \cdot P(\vec{\pi}_i|l')} \cdot P(x_i|\vec{\pi}_i, l) \\
&= \frac{P(l_c) \cdot P(\vec{\pi}_i|l_c) \cdot P(x_i|\vec{\pi}_i, l_c)}{\sum_{l'} P(l') \cdot P(\vec{\pi}_i|l')}
\end{aligned}$$

where the summation over $l$ collapses to a single leaf $l_c$ consistent with both $x_i$ and $\vec{\pi}_i$, since all other leaves $l$ have either $P(\vec{\pi}_i|l)$ or $P(x_i|\vec{\pi}_i, l)$ equal to zero. This equation gives us a simple way of calculating conditional distributions $P(X_i|\vec{\Pi}_i)$ from trees modeling joint distributions $P(X_i, \vec{\Pi}_i)$, assuming the distribution $P(X_i, \vec{\Pi}_i|l)$ within each leaf $l$ can be marginalized to compute $P(\vec{\Pi}_i|L)$ and conditionalized to compute $P(X_i|\vec{\Pi}_i, L)$.

The algorithm we use to learn joint density trees of this form is identical to the learning algorithm we use for decision / regression trees, except the joint density tree learning algorithm treats $X_i$ and $\vec{\Pi}$ on equal footing: either $X_i$ or a variable in $\vec{\Pi}_i$ can be tested at any particular branch node, and the joint log-likelihoods of all the variables in $\{X_i\} \cup \vec{\Pi}_i$ are used for evaluating any particular branch choice rather than just the conditional log-likelihood of $X_i$.

Joint density trees are trivially capable of representing Bayesian classifiers when used conditionally in this manner. In particular, since each leaf in the density trees employed in this paper models each discrete variable independently of all other variables (using a multinomial distribution), a Naive Bayes classifier for discrete variables is obtained in the special case where the density tree is a one-level density stump with a root node branching on the variable to be predicted. Such Naive Bayes classifiers have previously been used to model the conditional distributions within Bayesian networks (Heckerman and Meek, 1997). A commonly used Bayesian classifier for continuous variables is to model each class distribution with a Gaussian; this classifier is obtained simply with a density stump branching on the class variable with leaves employing Gaussian distributions over the continuous variables. More generally, suppose a joint density tree over discrete variables has a branch structure similar to the branch structure of a stratified conditional density tree: that is, once the output variable is tested in a branch node, no further tests can be performed on the input variables in subsequent levels of the tree. When this joint density tree is used to estimate conditional distributions for the output variable, it is similar in form and function to a hybrid decision tree / Naive Bayesian classifier also developed in previous research (Kohavi, 1996). In the most general case when the tree has an arbitrary branch structure (and the variables are not necessarily discrete), the algorithm for computing conditional distributions essentially creates a Bayesian classifier "on the fly" across different parts of the tree to determine which of the leaves consistent with $\vec{\pi}_i$ the datapoint probably came from.

Somewhat surprisingly, our experimental results show that learning joint density trees and then using them conditionally in this manner frequently leads to more accurate conditional density estimation than the more direct approach of learning and using stratified conditional density trees. One possible explanation of this phenomenon is discussed briefly at the end of the next section.

### 2.4 Approximately conditionalized joint density trees

The joint density trees discussed in the previous section can be learned quickly and they appear to be at least as accurate as stratified conditional density trees in our experiments. However, they are computationally expensive to use, since evaluating the denominator $\sum_{l'} P(l') \cdot P(\vec{\pi}_i|l')$ requires traversing the tree finding all leaves consistent with the known value of $\vec{\pi}_i$.

If the class of density functions used in the leaves is closed under addition and scalar multiplication, then we can take a density tree modeling $P(X_i, \vec{\Pi}_i)$ and precompute a *marginalized* density tree $P(\vec{\Pi}_i)$. Such a marginalization algorithm for density trees with constant-density leaves has been used in previous work by Kozlov and Koller on message-passing algorithms for inference in continuous-variable graphical models (Kozlov and Koller, 1997). Once this tree is computed, we can compute the conditional distribution simply as $P(X_i|\vec{\Pi}_i) = \frac{P(X_i, \vec{\Pi}_i)}{P(\vec{\Pi}_i)}$, where computing the numerator and computing the denominator each require locating and evaluating only one leaf distribution in the appropriate tree. Unfortunately, many useful leaf density estimators are not closed under addition, including those that have factored nonuniform distributions over multiple variables. Marginalizing trees with such leaves results in a marginalized tree whose leaves contain mixture distributions with many components, and evaluating these leaves can take a significant amount of computational time. Furthermore, for some operations we might wish to perform with density trees, such as sampling or compression, being able to compute $P(X_i|\vec{\Pi}_i)$ as a quotient of two black-box functions is not particularly helpful. Such operations are much more naturally performed in terms of leaf probabilities $P(L|\vec{\Pi}_i)$ and leaf-dependent conditional probabilities $P(X_i|L, \vec{\Pi}_i)$. (For example, suppose we have an algorithm capable of generating a random sample from a Gaussian distribution. It is simple to use this routine to generate a random sample from a mixture of Gaussians – first, we randomly choose the mixture component, and then we generate a random sample from the corresponding Gaussian distribution. On the other hand, it is not so straightforward to use it to generate a random sample from a distribution represented as a quotient of two Gaussian mixtures.)

However, in such situations we can still speed up the evaluation of conditional probabilities by creating an auxiliary tree in which each leaf contains a pointer to a single leaf of the original density tree. This auxiliary tree has the same structure as a stratified conditional density tree in that all branching on the parent variables is performed first, after which all branching is on the child variable. We create the auxiliary tree's structure by first using a marginalization algorithm similar to that employed by Kozlov and Koller. This marginalization algorithm produces a tree in which all branches over $X_i$ have been removed, and which contains one leaf for every distinct possible combination of leaves in the original tree that can be consistent with any single $\vec{\pi}_i$. We then recursively refine each leaf of the resulting marginalized tree by branching on $X_i$ until each of the resulting leaves has a nonzero intersection with precisely one of the leaves in the original joint density tree. See Figure 3 for an example; see the full paper (Davies, 2002) for further detail. We call the combination of the original joint density tree and the auxiliary tree a *conditionalized joint density tree*.

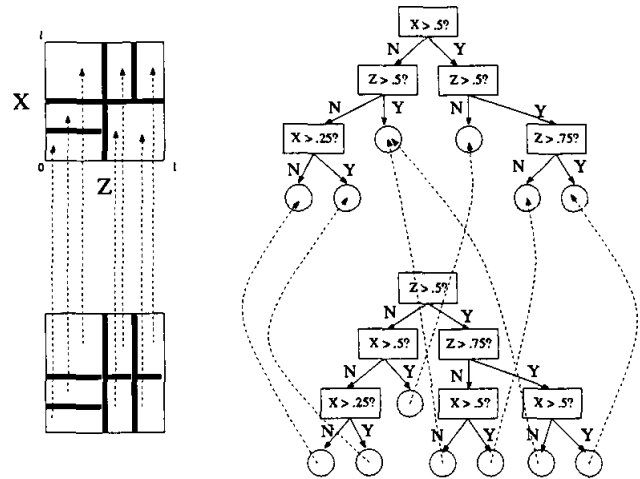

Figure 3: An example of a conditionalized joint density tree. Each leaf of the auxiliary tree (bottom) contains a pointer back up to a single leaf in the original density tree (top). The geometrical representation of each tree is shown to the left, and the tree-based representation to the right.

The auxiliary tree can help speed up the evaluation of conditional probabilities simply by providing (in a single, relatively small subtree) pointers to all the leaves in the original joint density tree that are consistent with any given value of $\vec{\Pi}_i$. This speedup is roughly similar to that which would be achieved by using a marginalized density tree with mixture models in the leaves to compute $P(\vec{\Pi}_i)$ — roughly a factor of two or so in our experiments (not shown here). However, we can speed up the conditional evaluation of joint density trees further by introducing a simple approximation. Within the context of any subtree $t_s$ where the branching on $X_i$ begins, we can approximate the conditional distribution $P(\vec{\Pi}_i|l')$ over each original density tree leaf $l'$ as $\hat{P}_s(\vec{\Pi}_i|l')$, the mean of $P(\vec{\Pi}_i|l')$ over all the datapoints consistent with $t_s$'s constraints. The conditional density can then be be computed approximately as:

$$\begin{aligned} P(x_i|\vec{\pi}_i) &\approx \frac{P(l_c) \cdot \hat{P}_s(\vec{\pi}_i|l_c)}{\sum_{l'} P(l') \cdot \hat{P}_s(\vec{\pi}_i|l')} \cdot P(x_i|\vec{\pi}_i, l_c) \\ &= \alpha_c^s P(x_i|\vec{\pi}_i, l_c) \end{aligned}$$

where $l_c$ is the single leaf consistent with both $x_i$ and $\vec{\pi}_i$

(as before) and $\alpha_c^s$ is a constant. (An alternative would be to compute $\alpha_c^s$ directly as the average of $P(l_c|\vec{\pi}_i)$ over the datapoints consistent with $t_s$'s constraints; however, this appears to not work quite as well empirically.) When we use a conditionalized joint density tree in this manner, we refer to it as an *approximately conditionalized joint density tree*.

If each leaf of the original joint tree employs a nonuniform distribution over the parent variables, then obtaining the conditional distribution $P(X_i|\vec{\pi}_i)$ from a joint tree using the relationship

$$P(x_i|\vec{\pi}_i) = \sum_l P(l|\vec{\pi}_i) \cdot P(x_i|\vec{\pi}_i, l)$$

can actually result in more accurate conditional density estimation than possible with stratified conditional density trees, even though the joint density trees are optimized for joint probabilities rather than conditional probabilities. Intuitively, by combining the distributions learned in different leaves using this relationship, we have essentially created a *"soft branch"* over $\vec{\Pi}_i$ that helps us to more accurately predict $X_i$ as a function of $\vec{\Pi}_i$ without actually splitting the dataset further into completely disjoint subsets. In fact, as our experimental results will show, conditionalized joint density trees can actually provide more accurate estimates than stratified conditional density trees even when the conditionalized joint density trees are used approximately, i.e., even when the "soft branching" coefficients $\alpha_c^s$ are fixed as constants.

### 2.5 Leaf types

In all of our experiments, each leaf represents the distribution of each discrete variable as a multinomial distribution independent of all other variables. However, we have experimented with a wide variety of distributions with which to represent the densities of continuous variables within each density tree leaf: constant (i.e. uniform) densities; Gaussians with diagonal covariance matrices or general covariance matrices (renormalized so our conditional distributions always integrate to one); exponential distributions; and linear and multilinear distributions. Space restrictions prevent us from discussing all of these possibilities here; see (Davies, 2002) for further detail. Of these, linear and multilinear interpolation appear empirically to be the best density approximators for use in the leaves of stratified conditional density trees and conditionalized joint density trees, with multilinear interpolation being slightly more accurate than linear interpolation but also more computationally expensive. With linear interpolation, each continuous variable is modeled independently, and its density varies linearly within each leaf. With multilinear interpolation, the $d$ continuous variables are modeled jointly by interpolating between $2^d$ densities associated with the corners of the leaf's bounding hyperbox. In both cases, each distribution to fit is expressed as a mixture model and then fit with the EM algorithm (Dempster et al., 1977) to maximize the log-likelihood of the training data. Because the distribution of each mixture component is fixed and only the prior probabilities of the mixture components are adjusted, EM can be performed relatively quickly — and the log-likelihood is convex, so there are no suboptimal local maxima for EM to get trapped in. In order to keep leaf-learning reasonably fast at the higher levels of the tree where many datapoints lie in each candidate leaf, we restrict the EM algorithm to using at most $25 * 2^d$ datapoints to fit any $d$-dimensional multilinear distribution, or at most $25 * 2 * d$ datapoints to fit any $d$-dimensional independent linear interpolation. Furthermore, we restrict EM to run for at most 10 iterations. Experiments not described in this paper have shown that this subsampling and this limitation on the number of iterations have a negligible effect on the accuracy of the resulting density estimator.

### 2.6 Smoothing

The tree-learning algorithms we employ are generally oriented towards maximizing the log-likelihood of the data — either just of $X_i$ in the case of CART-like and stratified conditional density trees, or of $\{X_i\} \cup \vec{\Pi}_i$ in the case of joint density trees. If we are using test-set log-likelihood as our criterion for density estimator quality, such maximum-likelihood estimates can perform arbitrarily poorly. Rather than attempt a complex fully Bayesian solution to the problem, we rely on a commonly used and simpler technique for working around it: namely, we adjust the overall distribution slightly towards the uniform distribution in an ad-hoc fashion. For simplicity, we assume some bounding box is known *a priori* for the continuous variables. See (Davies, 2002) for details and a discussion of how to handle other scenarios.

### 2.7 Experimental results

In this section we compare the accuracy of the four tree types described above on a simple synthetic dataset and on two large scientific datasets. The "Connected" synthetic dataset was generated by sampling 80,000 datapoints from a mixture of Gaussians in two dimensions. The "Bio" dataset contains data from a high-throughput biological cell assay. There are 12,671 records and 31 variables. 26 of the variables are continuous; the other five are discrete. Each discrete variable can take on either two or three different possible values. The "Astro" dataset contains data taken from the Sloan Digital Sky Survey, an extensive astronomical survey currently in progress. This dataset contains 111,456 records and 68 variables. 65 of the variables are continuous; the other three are discrete, with arities ranging from three to 81. See the full version of the paper (Davies, 2002) for further experiments on additional synthetic datasets, on other modifications of the scientific datasets, and with many other variations of the learning algorithms.

Two minor adjustments were made to each of the scientific datasets before handing them to any of our learning algorithms. First, all continuous variables were scaled so that all values lie within $[0, 1]$. This helps put the log-likelihoods we report in context, and possibly helps prevent problems with limited machine floating-point representation. Second, the value of each continuous value in the dataset were randomly perturbed by adding noise to it — either uniform noise with a range of .001, or Gaussian noise with a standard deviation of .001. This noise was added to eliminate any deterministic relationships or delta functions in the data. The log-likelihood of a continuous dataset exhibiting even a single deterministic relationship between two variables is infinite when given the correct model; in such a situation, it is not clear how meaningful log-likelihood comparisons between competing learning algorithms would be. Adding two different kinds of noise also allows us to check how sensitive the algorithms' relative performances are to variations in the small-scale details of the datasets.

Figure 4 shows a sample of our experimental results for CART-like vs. stratified conditional density trees. Two different kinds of leaf types are shown for the CART-like trees: Gaussians with constant means, and Gaussians in which the mean is a linear function of the parent variables as determined by linear regression. For stratified conditional density trees we show results for uniform-density leaves in addition to these two previous leaf types. A 10-fold cross-validation is performed; we show the mean of the log-likelihoods of the test sets, as well as its empirically estimated 95% confidence interval. The best algorithm for a given dataset is shown in bold italics, as well as all others that are not worse than it with at least 95% confidence according to a Student's t-test. In the case of the synthetic "Connected" dataset, the task is simply to model the conditional distribution of one variable given the other; in the case of the two scientific datasets, the task is to model the joint distribution over all the variables using a Bayesian network with a fixed structure. (These structures had been learned automatically in previous work (Davies and Moore, 2000)). The results show that stratified conditional density trees model the distributions much more accu-rately. This is unsurprising in the case of the synthetic dataset, which was generated to have multimodal conditional distributions; however, it is interesting to note that the scientific datasets also contain complex conditional distributions not adequately captured by CART-like conditional density trees. Unfortunately, stratified conditional density trees are also much more computationally expensive to learn; in the case of the Astro dataset, the experiment for stratified conditional density trees employing linear regression in the leaves was aborted because it would have taken several CPU-days to complete. (We omit the results for the scientific datasets with uniform noise added rather than Gaussian; however, they are qualitatively similar.)

Figure 4: Accuracies and learning times for CART-like vs. stratified density trees.

Figure 5 shows some of our experimental results comparing stratified conditional density trees vs. various forms of joint density trees. Two different leaf distributions are shown for each tree type: uniform, and independent linear interpolations for each variable ("ILI"). These results illustrate several important trends:

- Using interpolating distributions within the leaves improves accuracy over uniform distributions.

- Joint density trees with interpolating leaves are more accurate than stratified conditional density trees, and are fast to learn. Unfortunately, they're much more expensive to evaluate.

- Approximately conditionalized joint density trees are *still* more accurate than stratified conditional density trees, but are much faster to learn and about as fast to evaluate. Thus, these approxi-mately conditionalized joint density trees combine the best features of the three other tree types.

Figure 5: Stratified vs. Joint vs. Approximately Conditionalized Joint density trees.

Figure 6: Automatically learned Bayesian networks with density trees vs. global mixture models learned by AutoClass.

Similar results are obtained for other synthetic datasets and for the scientific datasets with Gaussian noise added (Davies, 2002).

### 2.8 Learning Bayesian Network Structures with Interpolating Conditional Density Trees

We have developed an iterative Bayesian network structure-learning algorithm capable of using different kinds of density trees for three different phases of the learning task. This algorithm is somewhat similar in spirit to the Sparse Candidate algorithm for learning network structures over discrete variables (Friedman et al., 1999), and can be seen as a heuristic approximation of steepest-ascent hill-climbing in order to make it computationally feasible. Up to three different kinds of density trees may be used for three different parts of the algorithm:

- Fast-to-learn but relatively inaccurate trees can be used to occasionally recompute "steepness" estimates in network structure search space, i.e., which arc additions and removals seem promising.

- Medium-quality trees can be used to compare a new candidate network structure with the best previously found network structure.

- Expensive, high-quality density trees can be used for the final network parameterization after a promising network structure has been settled upon.

Space restrictions preclude a detailed description of this network structure-learning algorithm; see (Davies, 2002) for details. Using this flexible network search algorithm allows us to learn Bayesian networks modeling joint probability distributions over many continuous and discrete variables in a reasonable amount of time. We compare the accuracy of these Bayesian networks with that of global mixture models learned over all variables simultaneously by AutoClass (Cheeseman and Stutz, 1996). The results in Figure 6 show that on the higher-dimensional "Astro" scientific dataset, our Bayesian networks provide significantly more accurate density estimation than the global mixture models, and can be learned and evaluated more quickly as well — even when Gaussian noise is added to the data, which would favor AutoClass's Gaussian mixture models. The difference on the Astro dataset is even more dramatic when the added noise is uniform. Which of the two approaches works better on the Bio dataset depends on the type of noise added; our networks fare better when the added noise is uniform, but AutoClass fares better when the noise is Gaussian.

## 3 CONCLUSIONS

We have explored a wide variety of tree-based representations for conditional density estimation, and shown that they can be used to feasibly learn Bayesian networks over dozens of continuous variables from many thousands of datapoints. In some cases, the resulting models are simultaneously more accurate, faster to learn, and faster to evaluate than global mixture models. We have not yet experimentally compared this approach to previously developed global discretization-based approaches to learn-ing Bayesian networks (e.g. (Friedman and Goldszmidt, 1996a), (Monti and Cooper, 1998a)); while we have presented an interesting possible alternative, further experimentation is warranted. Numerous other lines of further research are possible; for example, explicit accuracy/computation tradeoffs can be explored for approximately conditionalized joint density trees. See (Davies, 2002) for further discussion of these and other issues.

## References

L. Breiman, J. Friedman, R. Olshen, and C. Stone. *Classification and Regression Trees*. Chapman & Hall, 1984.

P. Cheeseman and J. Stutz. Bayesian classification (AutoClass): Theory and results. In U. M. Fayyad, G. Piatetsky-Shapiro, P. Smyth, and R. Uthurasamy, editors, *Advances in Knowledge Discovery and Data Mining*. MIT Press, 1996.

S. Davies. Fast Factored Density Estimation and Compression with Bayesian Networks. Ph.D. Thesis, Carnegie Mellon University, 2002.

S. Davies and A. Moore. Mix-nets: Factored Mixtures of Gaussians in Bayesian Networks with Mixed Continuous and Discrete Variables. In *Proceedings of the Sixteenth Conference on Uncertainty in Artificial Intelligence (UAI2000)*, 2000.

A. P. Dempster, N. M. Laird, and D. B. Rubin. Maximum likelihood from incomplete data via the EM algorithm. *Journal of the Royal Statistical Society*, B 39:1–39, 1977.

E. Driver and D. Morrell. Implementation of Continous Bayesian Networks Using Sums of Weighted Gaussians. In *Proceedings of the Eleventh Conference on Uncertainty in Artificial Intelligence (UAI95)*, 1995.

N. Friedman and M. Goldszmidt. Discretizing Continuous Attributes While Learning Bayesian Networks. In *Proceedings of the Thirteenth International Conference on Machine Learning*, pages 157–165, 1996a.

N. Friedman and M. Goldszmidt. Learning Bayesian Networks with Local Structure. In *Proceedings of the Twelfth Conference on Uncertainty in Artificial Intelligence (UAI96)*, 1996b.

N. Friedman and I. Nachman. Gaussian Process Networks. In *Proceedings of the Sixteenth Conference on Uncertainty in Artificial Intelligence (UAI2000)*, 2000.

N. Friedman, I. Nachman, and D. Peér. Learning Bayesian Network Structures from Massive Datasets: The Sparse Candidate Algorithm. In *Proceedings of the Fifteenth Conference on Uncertainty in Artificial Intelligence (UAI99)*, pages 206–215, 1999.

D. Heckerman and D. Geiger. Learning Bayesian networks: a unification for discrete and Gaussian domains. In *Proceedings of the Eleventh Conference on Uncertainty in Artificial Intelligence (UAI95)*, 1995.

David Heckerman and Christopher Meek. Embedded Bayesian network classifiers. Technical Report MSR-TR-97-06, Microsoft Research, Redmond, WA, March 1997.

R. Hofmann and V. Tresp. Discovering Structure in Continuous Variables Using Bayesian Networks. In D. S. Touretzsky, M. C. Mozer, and M. Hasselmo, editors, *Advances in Neural Information Processing Systems 8*. MIT Press, 1995.

R. Kohavi. Scaling Up the Accuracy of Naive-Bayes Classifiers: a Decision-Tree Hybrid. In *Proceedings of the Second International Conference on Knowledge Discovery and Data Mining (KDD-96)*, 1996.

A. Kozlov and D. Koller. Nonuniform dynamic discretization in hybrid networks. In *Proceedings of the Thirteenth Conference on Uncertainty in Artificial Intelligence (UAI97)*, 1997.

S. Monti and G. F. Cooper. A Multivariate Discretization Method for Learning Bayesian Networks from Mixed Data. In *Proceedings of the Fourteenth Conference on Uncertainty in Artificial Intelligence (UAI98)*, 1998a.

S. Monti and G. F. Cooper. Learning Hybrid Bayesian Networks from Data. In M. I. Jordan, editor, *Learning in Graphical Models*. Kluwer Academic Publishers, 1998b.

S. Monti and G. F. Cooper. A Latent Variable Model for Multivariate Discretization. In *Proceedings of the Seventh International Workshop on AI & Statistics (Uncertainty 99)*, 1999.

J. R. Quinlan. Induction of decision trees. *Machine Learning*, 1:81–106, 1986.